\documentclass[preprint,12pt]{elsarticle}

\usepackage{amssymb,algorithm,algorithmic,graphicx,color}
\usepackage{subfigure}
\usepackage{amssymb}
\usepackage{graphicx}

\usepackage{url}
\urldef{\mailsa}\path|{rogodi,majiro,belenmg}@us.es|
\newcommand{\Z}{{\mathbb Z}}
\newcommand{\proof}{\noindent {\bf Proof}}
\newcommand{\scst}{\scriptscriptstyle}

\newtheorem{definition}{Definition}[section]
\newtheorem{lemma}{Lemma}[section]
\newtheorem{theorem}{Theorem}[section]

\newtheorem{example}[lemma]{Example}

\begin{document}
\begin{frontmatter}
\title{Cubical Cohomology Ring of 3D Pictures
}


%
%
\author{R. Gonzalez-Diaz}
\ead{rogodi@us.es}
\ead[url]{http://www.personal.us.es/rogodi}
\author{M.J. Jimenez}
\ead{majiro@us.es}
\ead[url]{http://ma1.eii.us.es/miembros/majiro}
\author{B. Medrano}
\ead{belenmg@us.es}
\ead[url]{http://ma1.eii.us.es/miembros/medrano}
\address{Depto. de Matem\'{a}tica Aplicada I, Escuela Superior de
Ingenier\'{\i}a Inform\'{a}tica, Universidad
de Sevilla, Avda. Reina Mercedes, s/n, 41012, Sevilla (Spain)}

%

\begin{abstract}
Cohomology and cohomology ring of three-dimensional (3D) objects are topological invariants that characterize holes and their relations. Cohomology ring has been traditionally computed on simplicial complexes. Nevertheless, cubical complexes deal directly with the voxels in 3D images, no additional triangulation is necessary, facilitating efficient algorithms for the computation of topological invariants in the image context. In this paper, we present formulas to directly compute the cohomology ring of 3D cubical complexes without making use of any additional triangulation. Starting from a cubical complex $Q$ that represents a 3D binary-valued digital picture whose foreground has one connected component, we compute first the cohomological information on the boundary of the object, $\partial Q$ by an incremental technique; then, using a face reduction algorithm, we compute it on the whole object; finally, applying the mentioned formulas, the cohomology ring is computed from such information.

\end{abstract}

\end{frontmatter}

\section{Introduction}

\begin{figure}
\centering
\includegraphics[width=9cm]{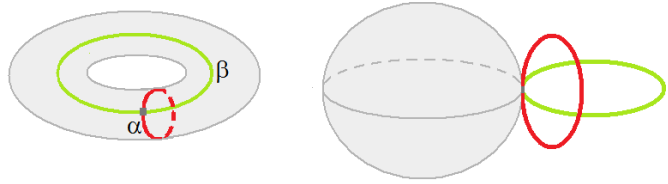}
\caption{On the left, a hollow torus and its two tunnels. On the right, the wedge sum of a  $2$-sphere and  two loops. }\label{ciclos}
\end{figure}

Many computer application areas involve topological methods which usually mean a significant reduction in the amount of data.
Homology is an algorithmically computable topological invariant that characterizes an object by its ``holes" (in any dimension). Informally, holes of a 3D-object are its connected components in dim. $0$, its tunnels in dim. $1$ and its cavities in dim. $2$. 
Cohomology is a topological invariant obtained by an algebraic duality of the notion of homology. Although the formal definition of cohomology is motivated primarily by algebraic considerations,  homology and cohomology of 3D objects are isomorphic, that is, they provide the same topological information. 
Nevertheless, cohomology has an additional ring structure provided by the cup product (denoted by $\smile$). The cup product can be seen as the way the holes obtained in homology are related to each other. For example, think of the torus, and the wedge sum  of two loops and a $2$-sphere. Both objects have two tunnels and one cavity; but the cavity ($\gamma$) of the first object can be decomposed in the product of the two tunnels ($\alpha$ and $\beta$), that is, $\alpha\smile\beta=\gamma$; the cavity of the second object cannot (see Fig. \ref{ciclos}). This information would contribute to a better understanding of the degree of topological complexity of the analyzed digital object, and would shed light on its geometric features. 

In \cite{GR03,GR05},
 a method for computing the cohomology ring of 3D
binary digital images is stated.
In those works, the cohomology ring
computation is performed over the (unique)
 simplicial complex associated
to the digital binary-valued picture using  the $14$-adjacency.
However,
one could assert that a more natural combinatorial structure when
dealing with 3D digital images is the one provided by cubical
complexes. 
One way to compute the cohomology ring  of a  cubical complex $Q$ is to convert it into a simplicial complex $K_Q$ by subdividing each cell and applying the known formulas for computing the cohomology ring of $K_Q$. In this paper, we present formulas to directly compute the cohomology ring of 3D cubical complexes without making use of additional triangulations. Besides, we describe a strategy to tackle the cohomology ring computation on a 3D binary-valued digital picture.  
This paper  extends a preliminary version (see \cite{GJM09}).

The paper is organized as follows. In Section \ref{polyhedral}, we recall the
concept of AT-model and extend it to general polyhedral cell
complexes; given the AT-model for a polyhedral cell complex, we give the formulas of a new AT-model obtained after  a subdivision; this result will be the key to prove the validity of the formulas of the cohomology ring of cubical complexes. In Section \ref{cupsection},
formulas for computing the cohomology ring of 3D cubical complexes are established.
Section \ref{main} is devoted to describe the process to obtain an AT-model of a 3D digital image that provides the ingredients for the cohomology ring computation. Finally, some conclusions and plans for future are
drawn in Section \ref{future}.

\section{AT-models for Polyhedral Cell Complexes}\label{polyhedral}

  In this section, we recall first the concept of AT-model (see \cite{GR03,GR05}) for a cell complex, which consists of an algebraic set of data that provides homological information. Given the AT-model for a polyhedral cell complex, we show the formulas of a new AT-model obtained after  a subdivision; this result will be the key for the formulas of the cohomology ring of cubical complexes in the next section.

Since we are working with objects embedded in $ {\mathbb R}^3$,
the homology groups are torsion-free \cite[ch.10]{AH35}, so computing homology over
a field is enough to characterize shapes \cite[p. 332]{Mun84}. This fact, together
with the isomorphism (over any field) between the homology and cohomology groups
\cite[p. 320]{Mun84}, enables us to consider ${\bf Z}/2$ as the ground ring
throughout the paper.

A {\it polyhedral cell complex} $P$ in $ {\mathbb R}^3$, is given by a finite
collection of cells which are convex polytopes (vertices, edges,
polygons and polyhedra), together with all their faces and such
that the intersection between two of them is either empty or a
face of each of them.  A {\it proper face} of $\sigma\in P$ is a
face of $\sigma$ whose dimension is strictly less than the one of $\sigma$.
A {\it facet} of $\sigma$ is a proper face of $\sigma$ of maximal
dimension. A {\it  maximal cell} of $P$ is a cell of $P$ which is not a proper face of any other cell of $P$.
Observe that  if the cells of $P$ are $n$-{\it simplices}, $P$ is a  {\it simplicial complex} (see \cite{Mun84}); in the case that the cells of $P$
are $n$-{\it cubes}, then $P$ is
a {\it cubical  complex} (see
\cite{KMM04}).
A $q$-cell of either a simplicial complex or a cubical complex can be denoted   by the list of its vertices.

For any graded set $S=\{S_q\}_q$, one can consider  formal sums
of elements of $S_q$, which are called $q$-{\it chains}, and which form
 abelian groups with respect to the component-wise
addition (mod $2$). These groups are called {\it $q$-chain groups} and denoted by $C_q(S)$. 
The collection of all
the chain groups associated to $S$ is denoted by ${\cal C}(S)=\{C_q(S)\}_q$ and called also chain group,
for simplicity. Let $\{s_1,\dots,s_m\}$ be the elements of $S_q$ for a fixed $q$.
Given two $q$-chains
$c_1=\sum_{i=1}^m\alpha_i s_i$ and $c_2=\sum_{i=1}^m\beta_i s_i$, where $\alpha_i,\beta_i \in {\bf Z}/2$ for $i=1,...,m$, the expression
 $\langle c_1, c_2\rangle$ refers to $\sum_{i=1}^m\alpha_i\cdot\beta_i\in {\bf Z}/2$.
For example, fixed $i$ and $j$, the expression $\langle c_1,s_i\rangle$ is $\alpha_i$   and $\langle s_i,s_j\rangle$ is $1$ if $i=j$ and $0$ otherwise.

 The {\it polyhedral chain complex} associated to the polyhedral cell complex $P$ is the
collection ${\cal C}(P)=\{C_q(P),
\partial_q\}_q$  where: 
\begin{itemize}
\item[(a)] each $C_q(P)$ is the corresponding chain group generated by the $q$-cells of $P$;
\item[(b)] the boundary operator $\partial_q :C_q(P)\rightarrow C_{q-1}(P)$ connects two immediate dimensions. The boundary of a $q$-cell is the formal sum of all its facets. It is extended to $q$-chains by linearity.
\end{itemize}
For example, consider a triangle $(v_i,v_j,v_k)$ with vertices $v_i$, $v_j$, $v_k$. The boundary of the triangle is the formal sum of its edges, that is, $\partial_2( v_i,v_j,v_k)=(v_i,v_j)+(v_j,v_k)+(v_i,v_k)$.

Given a polyhedral cell complex $P$, an algebraic-topological
model ({\it AT-model} \cite{GR03,GR05}) for $P$ is a set of data $(P,H,f,g,\phi)$,
where  $H$ is a graded subset of $P$ and $f,\,g,\,\phi$ are three families of maps $\{f_q:C_q(P)\rightarrow C_q(H)\}_q$,
$\{g_q:C_q(H)\rightarrow C_q(P)\}_q$ and
$\{\phi_q:C_q(P)\rightarrow C_{q+1}(P)\}_q$, such that, for each
$q$: 
\begin{itemize}
\item[(1)] $f_q g_q=id_{C_q(H)}$,  $\phi_{q-1}
\partial_q +
\partial_{q+1} \phi_q = id_{C_q(P)} + g_q f_q$, $f_{q-1}\partial_q=0$, $\partial_q g_q=0$;
\item[(2)] $\phi_{q+1}\,\phi_q=0$, $f_{q+1}\,\phi_q=0$,
$\phi_q\,g_q=0$.
\end{itemize}
 As a result, the chain group  ${\cal C}(H)$ is isomorphic to the homology (and to the cohomology) of
 $P$.
 In particular, the number of vertices of $H$ coincides with the number of connected components of $P$, the number of
edges of $H$ with the number of tunnels of $P$ and the number of $2$-cells of $H$ with
the number of cavities of $P$. Fixed $q$, for each $\sigma\in H_q$, $g_q(\sigma)$ is a representative cycle of a homology generator of dim. $q$. 
Define an homomorphism $\sigma^*f_q: C_q(P)\to {\bf Z}/2$ such that 
if $\mu$ is a $q$-cell of $P$ then 
$$\sigma^*f_q(\mu):=\langle \sigma, f_q(\mu)\rangle\;\;\mbox{  mod $2$}.$$
 Then, $\sigma^* f_q$ is a representative cocycle of a cohomology generator of dim. $q$. An isomorphism between ${\cal C}(H)$ and the homology (resp. cohomology) of $P$ maps each $\sigma\in H$ to the homology class represented by $g_q(\sigma)$ (resp. the cohomology class represented by $\sigma^*f_q$) (see \cite{GR03,GR05}).

\begin{figure}
\centering
\includegraphics[width=12cm]{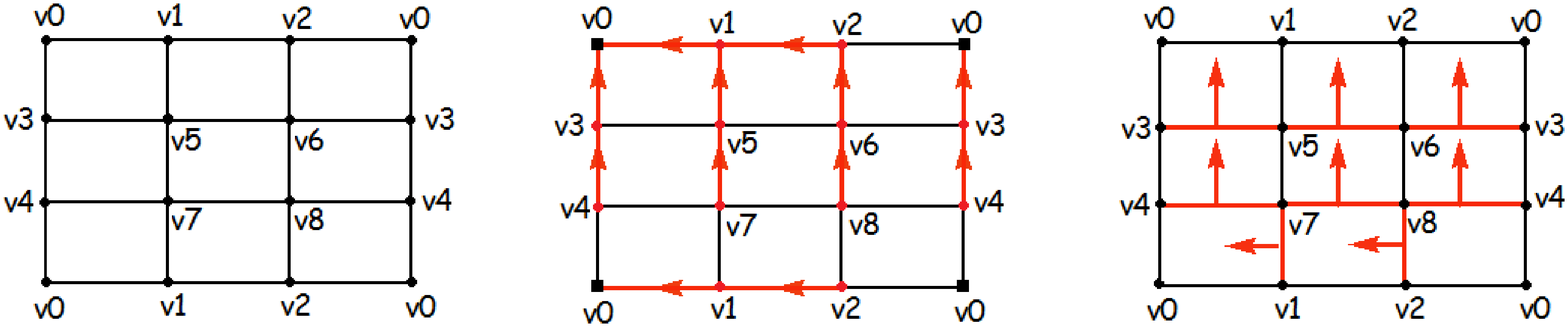}\\
\begin{footnotesize}a) \hspace{3.9cm} b)\hspace{3.9cm}c)$\mbox{ }$\hspace{2cm}\end{footnotesize}
\caption{a) An abstract cubical representation $Q$ of the hollow torus; b) the paths $\gamma_{(v_i,v_0)}$; c) the ``path" $c_e$, for any edge $e$ of $Q$.}\label{torocubico}
\end{figure}

 From now on, we will omit subscripts on behalf of simplicity.

\begin{example}\label{atmodeltoroesfera}
Let $Q$ be an abstract cubical representation of the hollow torus given in
Fig. \ref{torocubico}. An AT-model for $Q$ is the set of data $(Q,
H, f, g, \phi)$ given in the following table:

\begin{center}\begin{tabular}{| c | c | c || c | c |}
\hline

$\;Q\;$ & $\;f\;$  & $\;\phi\;$ & $\;H\;$ & $\;g\;$\\
\hline\hline
$v_0$ & $v_0$  & $0$ & $v_0$& $v_0$\\
$v_i$, $i=1, \ldots, 8$ & $v_0$ & $\gamma_{(v_i,v_0)}$&& \\
\hline
$a_i $, $i=1,2$ & $b_i$ & $c_{a_i}$&&  \\
$b_i$, $i=1,2$ & $b_i$  & $0$& $b_i$& $\alpha_i$\\
 any edge $b\neq a_i,b_i$ & $0$ & $c_b$&&  \\
\hline
 $c$ & $c$  & $0$& $c$& $\beta$\\
\,\, any square $\sigma\neq c$ \,& $0$& $0$& & \\
\hline
\end{tabular}\end{center}
where 
$a_1 \in \{( v_3, v_6), ( v_4, v_8)\}$; $a_2 \in \{( v_1, v_7), (v_2, v_8)\}$;
$b_1=(v_0,v_2)$; $b_2=(v_0,v_4)$; $c=(v_0,v_2,v_4,v_8)$;
$\gamma_{(v_i,v_0)}$ is the only path  in $Q$ from $v_i$ to
$v_0$ in  Fig. \ref{torocubico}.b (for example,
$\gamma_{(v_7,v_0)}=( v_5, v_7) +( v_1, v_5) + ( v_0, v_1)$); given an edge $e$ of $Q$, $c_e$ is 
the sum of the squares 
that correspond to the path starting from $e$ and following the arrows in Fig. \ref{torocubico}.c (for example,
$c_{( v_1,v_7)} = (v_0, v_1, v_4, v_7) +( v_3, v_5, v_4, v_7)+( v_0, v_1,
v_3, v_5)$); representative cycles of homology generators are  the vertex $v_0$, the tunnels 
$\alpha_1 = (v_0, v_1) + (v_1, v_2) + (v_0, v_2)$ and $\alpha_2 = (v_0, v_3) + (v_3, v_4) + (v_0,
v_4)$ and the cavity $\beta$ which is the sum of the $9$ squares of $Q$.
\end{example}

An algorithm for computing AT-models for polyhedral cell complexes appears for example in \cite{GR03,GR05}. 
In fact, in those papers, the algorithm is designed for simplicial complexes but the adaptation  to polyhedral cell complexes is straightforward. That algorithm runs in time $O(m^3)$ where $m$ is the number of cells of the given polyhedral cell complex.
If an AT-model $(P,H,f,g,\phi)$ for a polyhedral cell complex $P$ is computed using that algorithm then it is also satisfied that
if $a\in H$ then $f(a)=a$ and $a\in g(a)$.

\begin{figure}
\centering
\includegraphics[width=5cm]{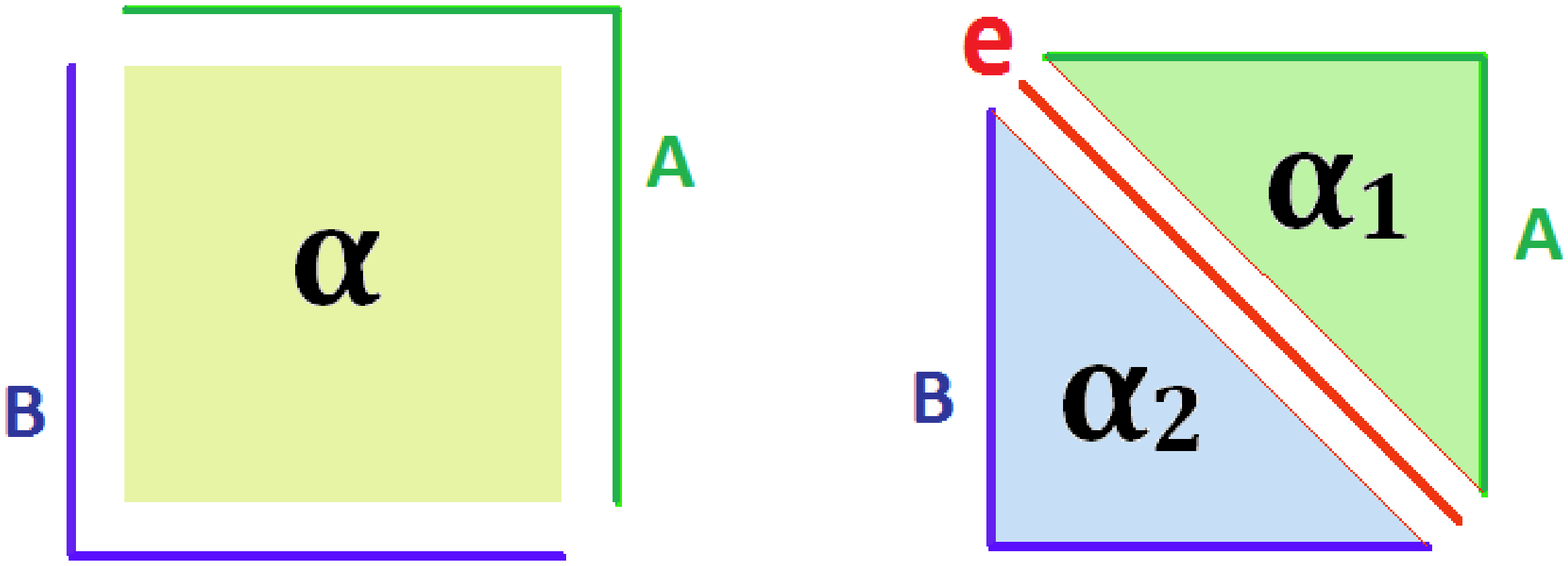}
 \caption{A
subdivision of a square $\alpha$ in two triangles $\alpha_1$ and $\alpha_2$.}\label{cuadrado}
\end{figure}

Let $P$ be a polyhedral cell complex. Fixed $q$, we say that a $q$-cell $\alpha\in P$ {\it is subdivided} into two new $q$-cells $\alpha_1\not\in P$ and $\alpha_2\not\in P$ by a new $(q-1)$-cell $e\not\in P$ (see, for example, Fig. \ref{cuadrado}) if: 
\begin{itemize}
\item[(a)] $e$ is a facet of $\alpha_1$ and $\alpha_2$;
\item[(b)] $\alpha_1\cup \alpha_2=\alpha$; 
\item[(c)] $\alpha_1\cap \alpha_2=e$.
\end{itemize}

The following lemma establishes how to obtain a new AT-model for $P$ after subdividing by a $(q-1)$-cell $e$ a $q$-cell $\alpha$ of $P$ into two new $q$-cells $\alpha_1$ and $\alpha_2$. This result will be used to prove the equivalence between the new formula to compute the cup product on cubical complexes and the classical one on simplicial complexes.

\begin{lemma}\label{subdivision}
Let 
$(P,H,f,g,\phi)$ be an AT-model for a polyhedral cell complex
$P$ computed using the algorithm given in \cite{GR03,GR05}. Let $\alpha$ be a $q$-cell,
 which is subdivided  into two $q$-cells $\alpha_1$ and $\alpha_2$ by
a new $(q-1)$-cell $e$. 
Let
$H':=(H\setminus\{\alpha\})\cup\{\alpha_1\}$ if $\alpha\in
H$, and  $H':=H$ otherwise;
$P':=(P\setminus\{\alpha\})\cup\{\alpha_1,\alpha_2, e\}$;
and $\partial'$ the boundary operator of $P'$
given by: $\partial'(c):=\partial(c)+\langle \alpha,
\partial(c)\rangle (\alpha + \alpha_1 + \alpha_2)$ for any $c \in P'\setminus \{e,\alpha_1,\alpha_2\}$.
Denote $\partial(\alpha_1)+e$ by $A$, and $\partial(\alpha_2)+e$ by $B$.
Then, the set $(P',H',f',g',\phi')$ is an AT-model for
$P'$, where $f'$, $g'$ and $\phi'$ are
given by:

\begin{itemize}
\item $f'(\alpha_1):=f(\alpha) + \langle \alpha, f(\alpha) \rangle (\alpha + \alpha_1)$;
$f'(\alpha_2):=0;\,f'(e):=f(A)(=f(B));$\\
$f'(\sigma):=f(\sigma) + \langle \alpha, f(\sigma)
\rangle (\alpha + \alpha_1)$, for any
$\sigma\in P'\setminus \{\alpha_1, \alpha_2,e\}$;

\item $\phi'(\alpha_1):=\phi(\alpha);\,\phi'(\alpha_2):=0;\,
\phi'(e):=\alpha_2+\phi(B)+\langle \alpha,
\phi(B)\rangle (\alpha+\alpha_1+\alpha_2);$\\
$\phi'(\sigma):=\phi(\sigma)+\langle \alpha,
\phi(\sigma)\rangle (\alpha+\alpha_1+\alpha_2)$, for any
$\sigma\in P'\setminus \{\alpha_1, \alpha_2,e\}$;

\item If $\alpha \in H$,
$g'(\alpha_1):=g(\alpha)+\alpha+\alpha_1+\alpha_2$;\\
$g'(\gamma):=g(\gamma)+\langle\alpha,g(\gamma)\rangle(\alpha+\alpha_1+\alpha_2)$,
for any $\gamma\in H'\setminus\{\alpha_1\}$.
\end{itemize}

\end{lemma}

\proof We have to check that $(P',H',f',g',\phi')$
is an AT-model for $P'$. 
We will 
only check that  $f'g'=id$ and $id + g'f' = \phi' \partial' +\partial'\phi'$. The rest of the conditions are left to the
reader.

Let $\gamma \in H'$, $\gamma\neq \alpha_1$, and $\sigma \in P' \backslash \{\alpha_1, \alpha_2, e\}$.

\begin{itemize}
\item $f' g' =id$:\\
 $f' g'(\gamma)=f'(g(\gamma) + \langle \alpha, g(\gamma) \rangle \alpha ) +\langle \alpha, g(\gamma) \rangle f'(\alpha_1)=fg(\gamma) + \langle \alpha, g(\gamma) \rangle f(\alpha ) +\langle \alpha, g(\gamma) \rangle \langle\alpha ,f(\alpha)\rangle(\alpha+\alpha_1)+\langle \alpha, g(\gamma) \rangle (f(\alpha )+\langle \alpha, f(\alpha) \rangle (\alpha+\alpha_1))=\gamma.$

 If $\alpha \in H$; $f'
g'(\alpha_1)=f'(g(\alpha)+\alpha) +
f'(\alpha_1)= f(g(\alpha)+ \alpha)+
f(\alpha) + \alpha + \alpha_1 =\alpha_1$.

\item  $id + g' f' = \phi' \partial' + \partial'\phi'$:\\
$\alpha_1 + g' f'(\alpha_1)=\alpha_1 + g'(f(\alpha)+\langle\alpha ,f(\alpha)\rangle\alpha)+\langle\alpha ,f(\alpha)\rangle g'(\alpha_1)
=\alpha_1 + g(f(\alpha)+\langle\alpha ,f(\alpha)\rangle\alpha)
+\langle\alpha ,g(f(\alpha)+\langle\alpha ,f(\alpha)\rangle\alpha)\rangle(\alpha+\alpha_1+\alpha_2)+\langle\alpha ,f(\alpha)\rangle g(\alpha)
+\langle\alpha ,f(\alpha)\rangle \\ (\alpha+\alpha_1+\alpha_2)
=\alpha_1 + gf(\alpha)+\langle\alpha ,gf(\alpha)+\langle\alpha ,f(\alpha)\rangle g(\alpha)\rangle(\alpha+\alpha_1+\alpha_2)
+\langle\alpha ,f(\alpha)\rangle(\alpha+\alpha_1+\alpha_2)
=\alpha_1 + gf(\alpha)+\langle\alpha ,gf(\alpha)\rangle (\alpha+\alpha_1+\alpha_2)
=\alpha_1 + \alpha+\phi(A)+\phi(B)+\partial\phi(\alpha)+\alpha+\alpha_1+\alpha_2
+\langle\alpha ,\phi\partial(\alpha)\rangle (\alpha+\alpha_1+\alpha_2)+\langle\alpha ,\partial\phi(\alpha)\rangle (\alpha+\alpha_1+\alpha_2)
=\alpha_2 +\phi(B)+\langle\alpha ,\phi(B)\rangle (\alpha+\alpha_1+\alpha_2)+\phi(A)
+\langle\alpha ,\phi(A)\rangle (\alpha+\alpha_1+\alpha_2)+\partial\phi(\alpha)+\langle\alpha ,\partial\phi(\alpha)\rangle (\alpha+\alpha_1+\alpha_2)
=\phi'\partial'(\alpha_1) +
\partial'\phi'(\alpha_1).$\\

$\phi'\partial'(\alpha_2) +
\partial'\phi'(\alpha_2)=\phi'(e)+\phi'(B)=\alpha_2 + \phi(B) + \langle \alpha,\phi(B)\rangle(\alpha +
\alpha_1 + \alpha_2) + \phi(B) + \langle
\alpha,\phi(B)\rangle(\alpha + \alpha_1 + \alpha_2)=\alpha_2=\alpha_2 + g'f'(\alpha_2)$.\\

 $\phi' \partial'(e) + \partial'\phi'(e)=
\phi'\partial(B) + \partial'(\alpha_2+
\phi(B)+\langle\alpha,\phi(B)\rangle(\alpha+\alpha_1+\alpha_2))=\phi\partial(B) + \partial(\alpha_2)+
\partial\phi(B) = e +gf(B)=e+
g'f'(e)$ (recall that $f(A)=f(B)=f(e)$ since $f\partial=0$ in an AT-model). \\

$\phi'\partial'(\sigma)+\partial'\phi'(\sigma)=\phi'(\partial (\sigma)+\langle\alpha,\partial(\sigma)\rangle\alpha) 
+\langle\alpha,\partial(\sigma)\rangle\phi'(\alpha_1+\alpha_2)+
\partial'(\phi (\sigma)+\langle\alpha,\phi(\sigma)\rangle\alpha)
+\langle\alpha,\phi(\sigma)\rangle\phi'(\alpha_1+\alpha_2)
=\phi(\partial (\sigma)+\langle\alpha,\partial(\sigma)\rangle\alpha) 
+\langle\alpha,\phi(\partial (\sigma)+\langle\alpha,\partial(\sigma)\rangle\alpha)\rangle(\alpha+\alpha_1+\alpha_2)
+\langle\alpha,\partial(\sigma)\rangle\phi(\alpha)+
\partial(\phi (\sigma)+\langle\alpha,\phi(\sigma)\rangle\alpha)
+\langle\alpha,\partial(\phi (\sigma)+\langle\alpha,\phi(\sigma)\rangle\alpha)\rangle(\alpha+\alpha_1+\alpha_2)+\langle\alpha,\phi(\sigma)\rangle\partial(\alpha)
=\sigma + gf(\sigma)+(\langle\alpha,\phi\partial(\sigma)\rangle+\langle\alpha,\partial\phi(\sigma)\rangle)(\alpha+\alpha_1+\alpha_2)
=\sigma + gf(\sigma)+\langle\alpha,\sigma+gf(\sigma)\rangle(\alpha+\alpha_1+\alpha_2).$\\

Besides, $\sigma + g'f'(\sigma)=\sigma+g'(f(\sigma)+\langle\alpha,f(\sigma)\rangle\alpha)+\langle\alpha,f(\sigma)\rangle g'(\alpha_1)
=\sigma+g(f(\sigma)+\langle\alpha,f(\sigma)\rangle\alpha)
+\langle\alpha,g(f(\sigma)+\langle\alpha,f(\sigma)\rangle\alpha)\rangle(\alpha+\alpha_1+\alpha_2)
+\langle\alpha,f(\sigma)\rangle(g(\alpha)+\alpha+\alpha_1+\alpha_2)
=\sigma+gf(\sigma)+\langle\alpha,gf(\sigma)\rangle(\alpha+\alpha_1+\alpha_2).$
\hfill{$\Box$}
\end{itemize}

Observe that $h: {\cal C}(H)\to $ ${\cal C}(H')$,  given
by $h(\alpha)=\alpha_1$ if $\alpha\in H$ and $h(\sigma)=\sigma$
for any $\sigma\in H\setminus\{\alpha\}$, is a chain-group isomorphism.

\section{3D Cubical Cohomology Ring}
\label{cupsection}

In \cite{vessel,KMM04}, the authors consider cubical complexes as the geometric building
blocks to compute the homology of digital images. In this section, we adapt to the cubical
setting, the method developed in \cite{GR03,GR05} for computing
the simplicial cohomology ring of 3D binary-valued digital
pictures. We must mention \cite{Ser51,Kad98} as related works
dealing with the cup product on cubical chain complexes in a theoretical context.

Since we are working with objects embedding in ${\mathbb R}^3$, it is satisfied that homology and cohomology are isomorphic. 
However,  cohomology
 has the advantage over homology of having an additional
ring structure given by the {\em cup product}, 
that is a topological invariant. This product provides information about the relationship between
the generators of (co)homology, that enables  to discriminate,
for instance, pairs of cycles in different contexts, as in Fig. \ref{ciclos}. 
Notice that, in 3D, the only non-trivial cup products are those
corresponding to elements of cohomology of dim. $1$. If the cup product of two elements of cohomology of dim. $1$ is not zero, then it is a sum of elements of cohomology of dim. $2$.
Recall that given an AT-model $(P,H,f,g,\phi)$ for a polyhedral cell complex $P$, it is satisfied that $H$ is isomorphic to the homology  and to the cohomology of $P$.

\subsection{Cohomology Ring of Simplicial Complexes}
We recall now how the cup product is defined in the simplicial setting using AT-models:

\begin{definition}\label{cupsimplicial}\cite{GR03,GR05}
Let $K$ be a simplicial complex. It is assumed that the vertices of $K$ are ordered. Let $(K,H,f,g,\phi)$ be an AT-model for $K$.
Let $\{\beta_1,\dots,{\beta_q}\}$ be the set of $2$-simplices of
$H$ and let $\alpha_1$ and $\alpha_2$ be two edges of $H$.  The {\it cup product} of  $\alpha_1$ and $\alpha_2$ is:
$$\sum_{k=1}^q ((\alpha_1 f \smile  \alpha_2 f)(g(\beta_k)))\beta_k \; \mbox{ mod } 2;$$
where $\alpha_1 f \smile  \alpha_2 f$ on a $2$-simplex
$(v_i,v_j,v_k)$  with vertices  $v_i<v_j<v_k$ is $\langle \alpha_1,
f(v_i,v_j)\rangle\cdot \langle \alpha_2, f(v_j,v_k)\rangle$; 
and $(\alpha_1 f \smile  \alpha_2 f)$ is extended to $2$-chains (sums of $2$-simplices) by linearity.
\end{definition}
Observe that for each $k$, $g(\beta_k)$ is a sum of $2$-simplices representing one cavity. Then, 
$(\alpha_1 f \smile  \alpha_2 f)(g(\beta_k))$ is a sum of $0$s and $1$s over ${\bf Z}/2$ whose result is $0$ or $1$. Therefore, 
$\sum_{k=1}^q ((\alpha_1 f \smile  \alpha_2 f)(g(\beta_k)))\beta_k$ is a sum of $2$-simplices  of $H$ of dim. $2$,
representing the cavities obtained by ``multiplying" the two representative cycles $g(\alpha_1)$ and $g(\alpha_2)$ (think of the two tunnels of a hollow torus).

It is known  that two objects with non-isomorphic cohomology rings, are not topologically equivalent (more precisely, they are not homotopic) 
\cite{Mun84}. 
To 
use the information of the cohomology ring for this aim, one can construct both matrices $M$ and $M'$ collecting the results of the cup product of cohomology classes of dim. $1$ of each object, if the rank of $M$ and $M'$ are different, then we can assert that both objects are not homotopic (see \cite{GR05}). 

\subsection{Cohomology Ring of Cubical Complexes}
Now, let $Q$ be a cubical complex. 
Our aim is to obtain a direct formula for  the cup product on $Q$ without making use of any triangulation.  

Suppose that the vertices of $Q$ are labeled in a way  that:
\begin{itemize}
\item[(P1)] \label{p1} Each square 
 $(v_i,v_j,v_k,v_{\ell})$ of $Q$ with vertices
$v_i<v_j<v_k<v_{\ell}$ has the edges $(v_i,v_j)$, $(v_i,v_k)$,
$(v_j,v_{\ell})$ and $(v_k,v_{\ell})$ in its boundary.
\end{itemize} 
For example, a cubical complex  whose set of vertices is a
subset of ${\mathbb Z}^3$ (the set of points with integer coordinates in 3D
space ${\mathbb R}^3$) with vertices labeled using  the
lexicographical order, satisfies P1.

\begin{figure}
\centering
\includegraphics[width=10cm]{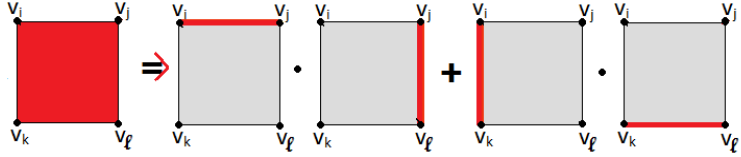}
 \caption{Scheme of the cubical cup product.}\label{ladoscuadrado}
\end{figure}

\begin{definition}\label{cupcubico} Let $Q$ be a cubical complex satisfying P1 and $(Q,H,f,g,\phi)$ an AT-model for $Q$.
Let $\{\beta_1,\dots,{\beta_q}\}$ be the set of squares of
$H$ and let $\alpha_1$ and $\alpha_2$ be two edges of $H$.  The {\it cup product} of 
$\alpha_1$ and $\alpha_2$ is
$$\sum_{k=1}^q ((\alpha_1 f \smile_{\scriptscriptstyle Q} \alpha_2 f)(g(\beta_k)))\beta_k\; \mbox{ mod } 2;$$
where $\alpha_1 f \smile_{\scriptscriptstyle Q}  \alpha_2 f$ on a square
$(v_i,v_j,v_k,v_{\ell})$  with vertices  $v_i<v_j<v_k<v_{\ell}$  (see Fig. \ref{ladoscuadrado}) is: 
$$\langle \alpha_1,
f(v_i,v_j)\rangle\cdot \langle \alpha_2, f(v_j,v_{\ell})\rangle+
\langle \alpha_1,
f(v_i,v_k)\rangle\cdot \langle \alpha_2, f(v_k,v_{\ell})\rangle;$$ 
 and $(\alpha_1 f \smile_{\scriptscriptstyle Q}  \alpha_2 f)$ is extended to $2$-chains (sums of square) by linearity.
\end{definition}

For simplicity, we sometimes use the notations $(\alpha \smile_{\scriptscriptstyle Q} \alpha' )(\beta)$ for $(\alpha f \smile_{\scriptscriptstyle Q}
\alpha' f)(g(\beta))$, and  analogously for the simplicial cup product.

\begin{example}\label{toroesfera}
Let $Q$ be an abstract cubical representation of the hollow torus given in
Fig. \ref{torocubico}. Consider the AT-model  $(Q,
H, f, g, \phi)$ for $Q$, given in Example \ref{atmodeltoroesfera}.
Recall that $H=\{v_0,(v_0,v_2),(v_0,v_4),(v_0,v_2,v_4,v_8)\}$; and
$g(v_0)=v_0$, $g(v_0,v_2)=(v_0,v_1)+(v_1,v_2)+(v_0,v_2)$, $g(v_0,v_4)=
(v_0,v_3)+(v_3,v_4)+(v_0,v_4)$ and $g(v_0,v_2,v_4,v_8)$ is the sum of the squares of $Q$,
representing the connected component, the two tunnels and the cavity.

Apply the formula given in Def. \ref{cupcubico} in order to obtain
the cup product of  $(v_0,v_2)$ and $(v_0,v_4)$ in $H$:
$$\begin{array}{l} 
((v_0,v_2)^* f \smile_{\scriptscriptstyle Q}  (v_0,v_4)^* f)(g( v_0,
v_2, v_4, v_8))\\
\hspace{2cm} :=\langle (v_0,v_2), f(v_0,v_2)\rangle \cdot
\langle (v_0,v_4), f(v_2,v_8) \rangle\\
\hspace{2.5cm} +
\langle (v_0,v_2), f(v_0,v_4)\rangle \cdot\langle (v_0,v_4),
f(v_4,v_8) \rangle\\
 \hspace{2cm}\,= 1 \cdot 1 + 0 \cdot 0 = 1.
\end{array}$$
then,
$(v_0,v_2) \smile_{\scriptscriptstyle Q} (v_0,v_4) = ( v_0, v_2, v_4,v_8)$. Recall that $( v_0, v_2, v_4,v_8)$ is the square in $H$ representing the cavity of the hollow torus. Therefore, the product of the two tunnels of the hollow torus is the cavity. 
  \end{example}

The following theorem shows the validity of the definition of $\smile_{\scriptscriptstyle Q}$  (Def. \ref{cupcubico}).
That is, it is stated that we obtain the same result by applying the formula of Def. \ref{cupcubico} to compute the cup product on the cubical complex, than making first a triangulation in order to obtain a simplicial complex, and applying the classical definition of the cup product given in Def. \ref{cupsimplicial}, afterwards. 

\begin{figure}
\centering
\includegraphics[width=11cm]{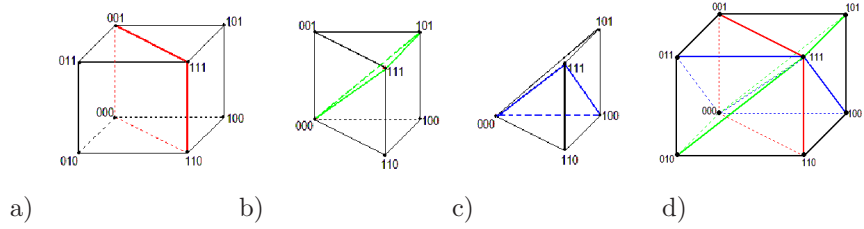}\\
\begin{footnotesize}a) \hspace{2.5cm} b)\hspace{2.5cm}c)\hspace{2.5cm}d)\hspace{3cm}$\mbox{ }$\end{footnotesize}
\caption{From a) to c), successive subdivisions; d) a cube
subdivided in $6$ tetrahedra.}\label{subdivisiones}
\end{figure}

Consider successive subdivisions of
each cube of a given cubical complex $Q$ until each one is converted in six tetrahedra,
and such that each square $(v_i,v_j,v_k,v_{\ell})$ of $Q$  is
subdivided by the edge $(v_i,v_{\ell})$ (see Fig. \ref{subdivisiones}.d). Let us
denote this resulting simplicial complex  by $K_{Q}$.  Observe
that with this particular subdivision, if $(v_p,v_q,v_r)$ is a
$2$-simplex of $K_{Q}$, with $v_p<v_q<v_r$, obtained by a
subdivision of a square of $Q$, then $(v_p,v_q)$ and $(v_q,
v_r)$ will correspond to edges in $Q$.

\begin{theorem}\label{equivalence}
Let $(Q,H,f,g,\phi)$ be an AT-model for $Q$. Let $\alpha$ and $\alpha'$ be two edges of $H$ and $\beta \in H$ a square.
Let $(K_Q,H',f',g',\phi')$ be the
AT-model for $K_{Q}$ obtained after successively applying Lemma \ref{subdivision}. Then,
$$(\alpha \smile_{\scriptscriptstyle Q} \alpha') (\beta) = (\alpha \smile \alpha')(h(\beta))$$
where $\smile$ is the simplicial cup product given in Def. \ref{cupsimplicial},
$\smile_{\scriptscriptstyle Q}$ is the cubical cup product given in Def. \ref{cupcubico},
 and
$h:{\cal C}(H)\to {\cal C}(H')$ is the isomorphism defined
at the end of Section \ref{polyhedral}.
\end{theorem}

\proof Observe that since $\alpha$ and $\alpha'$ are edges of $H$, then $\alpha, \alpha' \in
H'$, that is, $h(\alpha)=\alpha$ and $h(\alpha')=\alpha'$.
Observe also that $f(a)=f'(a)$ on any edge $a$ since $\beta \notin f(a)$.

We have to prove that
$(\alpha f \smile_{\scriptscriptstyle Q} \alpha' f)(g(\beta)) = (\alpha f' \smile
\alpha' f')(g'(h(\beta)))$.

Let $\beta=( v_i, v_j, v_k, v_{\ell})$. Let $\beta_1=( v_i, v_j, v_{\ell})$ and $\beta_2=(v_i,
v_k, v_{\ell})$ be the two triangles obtained after subdividing
$\beta$ by the edge $e=(v_i, v_{\ell})$. Remember that
$h(\beta)=\beta_1$ and $g'(h(\beta))=g'(\beta_1)$. Notice that $g'(\beta_1)$ coincides with $g(\beta)$ if we replace $\beta$ by $\beta_1+\beta_2$ in the expression of $g(\beta)$. 
 Therefore, it is enough to prove that $(\alpha f
\smile_{\scriptscriptstyle Q} \alpha' f)(\beta)=(\alpha f' \smile \alpha' f')(\beta_1 + \beta_2)$:
$$\begin{array}{l} (\alpha f \smile_{\scriptscriptstyle Q} \alpha' f)(\beta)=(\alpha f \smile_{\scriptscriptstyle Q} \alpha' f)( v_i, v_j, v_k, v_{\ell})\\
= \langle \alpha, f( v_i, v_j) \rangle \cdot
\langle \alpha', f( v_j, v_{\ell})\rangle + \langle \alpha,
f( v_i, v_k ) \rangle\cdot
\langle \alpha', f(v_k, v_{\ell}) \rangle\\
=\langle \alpha,
f'(v_i, v_j) \rangle \cdot \langle \alpha',
f'(v_j, v_{\ell}) \rangle+ \langle \alpha,
f'( v_i, v_k)\rangle \cdot \langle \alpha',
f'(v_k, v_{\ell})\rangle \\
  =(\alpha f' \smile \alpha' f')(\beta_1 + \beta_2) .
\end{array}$$
This concludes the proof. \hfill{$\Box$}

\section{Cubical Cohomology Ring of 3D Digital Pictures}\label{main}

In this section, we develop the main bulk of the paper:
beginning from a cubical complex $Q$ that represents a 3D binary
digital picture whose foreground has one connected component, first we compute an AT-model for
the  boundary $\partial Q$ of the object; then, having in mind that the homology of $\partial
Q$ contains the homology of $Q$, we obtain an AT-model for $Q$ with the representative cycles of
homology generators lying in $\partial Q$;
finally, applying the formula given in Section \ref{cupsection}, the
cohomology ring is computed from such an AT-model.

\subsection{From  Digital Pictures to Cubical Complexes}\label{picture}
 Each point of $\mathbb{Z}^3$ can be identified with a  unit cube (called {\it voxel}) centered at this point, with facets parallel to the coordinate planes.
This gives us an intuitive
and simple correspondence between points in $\Z^3$  and voxels in $\mathbb{R}^3$.

Consider a {\it 3D binary digital picture}
$I=(\Z^3,26,6, B)$, where $B$ (the {\it foreground}) is finite, having  $\Z^3$ as the
underlying grid and fixing the  $26$-adjacency for the points of $B$ and the $6$-adjacency for the points of $\mathbb{Z}^3\setminus B$ (the {\it background}).
We say that a voxel $V$ is in the {\em boundary} of $I$ if $V\in B$ (i.e., if the point of $\Z^3$ identified with $V$ is in $B$) and $V$ has a $6$-neighbor in $\Z^3\setminus B$. 

Take the  cubical complex $Q$ for $B$  whose
elements are the unit cubes (voxels) centered at the points of $B$ together with all their faces.
 Observe that this cubical complex with vertices labeled by the corresponding cartesian coordinates and considering the lexicographical order, satisfies (P1) (see page \pageref{p1}). 
 
 Without lack of generality, we consider that the foreground is connected.
 
The elements of  $\partial Q$ are  
 all the squares of $Q$ which are
shared by a voxel of $B$ and a voxel of $\Z^3\setminus B$ together with all their faces.

\begin{figure}
\centering
\includegraphics[width=8cm]{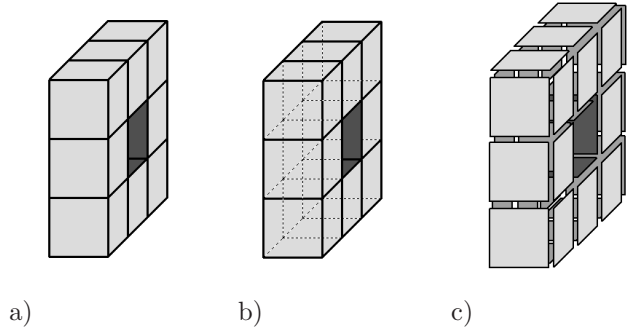}\\
\begin{footnotesize} a) \hspace{2.5cm} b)\hspace{2.5cm}c)\hspace{2.5cm}$\mbox{ }$\end{footnotesize}
\caption{a) A $3D$ binary digital picture $I=(\Z^3,26,6,B)$; b) the cubical
complex $Q$ for $B$; c) the set of squares of $\partial Q$.
}
\end{figure}

\subsection{AT-model for  $\partial Q$}\label{partialq}

Our interest now is to adapt the incremental algorithm for computing an AT-model given in \cite{GR03,GR05} to the particular complex
$\partial Q$.

First, 
consider the set of edges and vertices of $\partial Q$ as a graph and
compute a spanning forest $T$. Let $T_1,\dots,T_m$ be the trees of $T$ corresponding to the connected components of $\partial Q$. Fixed $i$, $i=1,\dots,m$, take a vertex $v_i$ of $T_i$ and consider it as the root of $T_i$.   

\begin{algorithm}\label{incremental} Computing an AT-model $(\partial Q,$ $H,$ $f,$ $g,$ $\phi)$ for $\partial Q$.
\end{algorithm}
\begin{tabbing}
{\sc Input:} \= {\tt The complex $\partial Q$,}\\
\> {\tt     the set $\{T_1,\dots, T_m\}$
 of trees of a spanning 
 forest $T$ of $\partial Q$, }\\
 \> {\tt the set  $\{v_1,\dots,v_m\}$ of roots of the trees of $T$.}\\
\end{tabbing}
\vspace{-1cm}
\begin{itemize}
\item[1.] \begin{tabbing}
 {\tt Initialize $f(\sigma):=\sigma,$ $\phi(\sigma):=0$ for any $\sigma\in \partial Q$; $H:=\{v_1,\dots,v_m\}$,}\\
   $U:=\{v:\; v \mbox{ \tt is a vertex of } \partial Q\}$,\\
    $f(v):=v_i \mbox{ \tt if $v$ is a vertex of } T_i$ for some $i$, $i=1,\dots,m$.\\
 {\tt For }\= {\tt $i=1$ to $m$ do}\\
 \> {\tt From }\= {\tt $\ell=1$ to the height of $T_i$ do}\\
 \>\> {\tt For }\= {\tt  each vertex $v$ at level $\ell$, and edge $a$ linking $v$}\\
 \>\> \>{\tt with   its  parent $w$ do}\\
 \>\>\> {\tt $\phi_i(v):=a+\phi_i(w)$, $U:=U\cup\{a\}$, $f(a):=0$.}
\end{tabbing}
\item[2.] \begin{tabbing}
{\tt  While there are  edges in $\partial Q\setminus U$ do}\\
 {\tt $\;$ If } \= {\tt there is a square $c\in \partial Q\setminus U$ with exactly one edge}\\
\> {\tt  $a\in \partial Q\setminus U$ in its boundary do $U:=U\cup\{c,a\}$, }\\
 \> {\tt  $f(a):= f(\partial(c)+a)$, $\phi(a):=c+\phi(\partial(c)+a)$, $f(c):=0$.} \\
 {\tt $\;$ Else }\= {\tt take an edge $a\in \partial Q\setminus U$ then $H:=H\cup\{a\}$, $U:=U\cup\{a\}$.}
\end{tabbing}
\item[3.] \begin{tabbing}
{\tt While  there is  a square $c$ in $\partial Q\setminus U$ do  $U:=U\cup \{c\}$.}\\
{\tt $\;$ If }\= {\tt $f\partial(c)=0$ then
 $H:=H\cup\{c\}$ .}\\
{\tt $\;$ Else  take an edge $a$ in $f\partial(c)$  then $H:=H\setminus\{a\}$.} \\
 {\tt $\;$ $\;$ For }\={\tt each edge $b$ in $\partial Q\setminus T$  do $f(b):= f(b)+ \langle a, f(b)\rangle f\partial(c)$,}\\
 \> $\phi(b):= \phi(b)+\langle a, f(b)\rangle(c+ \phi\partial(c))$, $f(c)=0$.
\end{tabbing}
\item[4.] \begin{tabbing}
 {\tt For  each $\sigma\in H$ do  $g(\sigma):=\sigma+\phi\partial(\sigma)$.}
 \end{tabbing}
 \end{itemize}
\begin{tabbing}
\vspace{-1cm}
{\sc Output:} {\tt the AT-model $(\partial Q, H,f,g,\phi)$ for $\partial Q$.}
\end{tabbing}

The auxiliary set $U$ is defined to indicate the cells which have already been used. In Step 1,
neither the vertices nor the edges of $T_i$ create cycles except for the root $v_i$. 
 In Step 3,
  if a square has edges in its boundary that created cycles in a previous step, then
one of these cycles is destroyed. 
Otherwise, this square creates a new cycle (a cavity). In the last step, the representative cycles of homology generators are computed.

Observe that all the steps of Alg. \ref{incremental} are quadratic in the number of elements of $\partial Q$ (worse case complexity) 
except for the last part of Step 3 which is cubic in the number of edges of $\partial Q\setminus T$.

\begin{figure}
\centering
\includegraphics[width=12cm]{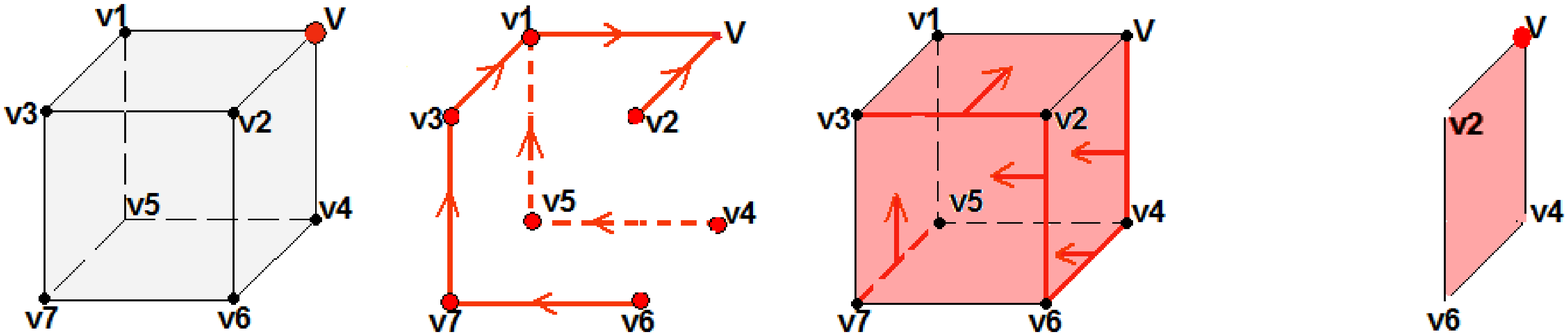}\\
\begin{footnotesize}a) \hspace{2.5cm} b)\hspace{2.5cm}c)\hspace{2.5cm}d)\hspace{2.5cm}$\mbox{ }$\end{footnotesize}
\caption{a) A hollow cube $\partial Q$; b) a spanning tree $T$ with root $v$; c) the ``paths'' $c_b$; d) the cells of $H$.}
\label{hollowcube}\end{figure}

\begin{example} The AT-model $(\partial Q,H,f,g,\phi)$ of a hollow cube $\partial Q$ (see Fig. \ref{hollowcube}) is:

\begin{center}\begin{tabular}{c|c|c|c||c|c|}
\hline
&$\partial Q$ &$f$ &  $\phi$&$H$ & $\;g\;$\\
\hline\hline
\mbox{Step 1} & $v$ & $v$ & $0$&$v$& $v$ \\
&  $v_i$, $i=1, \ldots, 7$ & $v$  & $\gamma_{(v_i,v)}$&&\\
\hline
\mbox{Step 2} & Any edge $b \in Q \backslash T$ & 0   & $c_{b}$&&\\
\hline
\mbox{Step 3} & $(v, v_2, v_4, v_6)$ & $(v, v_2, v_4, v_6)$  & $0$&$(v, v_2, v_4, v_6)$& $C$\\
\hline
\end{tabular}\end{center}
where $\gamma_{(v_i,v)}$ is the only path  in $T$ from $v_i$ to
$v$;
$c_b$ is the
square  at which the arrow corresponding
to an edge $b$ in Fig. \ref{hollowcube}.c points, except for $c_{(v_4, v_6)}$ which is
$( v_4, v_5, v_6, v_7) + (v_1, v_3, v_5, v_7)$; and $C$ is the sum of the six squares of $\partial Q$.
\end{example}

\subsection{AT-model for $Q$}\label{q}

Now, we  use  a {\it face reduction} technique 
(see, for example, \cite{GJMMR08,KMM04,PIKDH08})
in order to obtain a cell complex $K$ such that
 homology, cohomology and cohomology ring of $K$ coincide with that of $Q$, and such that the cells of $\partial Q$ are also cells of $K$. 

\begin{algorithm} Face Reduction Process.

\begin{tabbing}
{\sc Input:} {\tt A cubical complex $Q$}.
{\tt Initially, $K:=Q$.}\\
{\tt While } \= {\tt there exist $\sigma,\sigma'\in Q\setminus \partial Q$ such that $\sigma'$ is in $\partial(\sigma)$ do}\\
\> {\tt For }\= {\tt each cell $c\in K$ such that $\sigma'$ is in $\partial(c)$ do}\\
\>\> {\tt redefine $\partial(c)$ as $\partial(c+\sigma)$.}\\
\> {\tt Remove $\sigma$ and $\sigma'$ from the current $K$;}\\
{\sc Output:} {\tt the cell complex $K$}. 
\end{tabbing}
\end{algorithm}

\begin{figure}
\centering
\includegraphics[width=10cm]{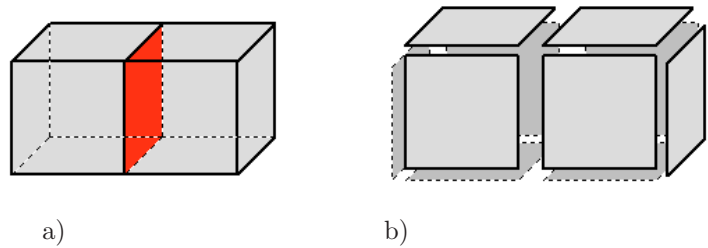}\\
\begin{footnotesize} a) \hspace{4cm} b)\hspace{3.5cm}$\mbox{ }$\end{footnotesize}
\caption{a) A cubical complex $Q$ composed by two cubes $c$ and $\sigma$ sharing a square $\sigma'$  (in red) and all their faces; 
 b) the squares of $\partial (c):=\partial(c+\sigma)$ in $Q'$.}
\end{figure}

Observe that after the face reduction process, we obtain a cell complex $K$ with the same topological information as $Q$ but with much less cells.  
Now, starting from an AT model for $\partial Q$, compute an AT-model for $K$ adding the cells of $K\setminus \partial Q$ incrementally 
as follows:

\begin{algorithm}\label{alghom} AT-model for $K$. 

\begin{tabbing}
{\sc Input:} \= {\tt An AT-model for $\partial Q$: $(\partial Q,H,f,g,\phi)$ and the cells }\\
 \> {\tt $\{\sigma_1,\dots,\sigma_m\}$ of $K\setminus \partial Q$ ordered by increasing dimension.}\\
{\tt Initially,} \=  {\tt  $f_{\scst K}(\sigma):=f(\sigma)$,
      $\phi_{\scst K}(\sigma):=\phi(\sigma)$, $g_{\scst K}(\sigma)=g(\sigma)$ for each $\sigma\in \partial Q$;}\\
      \> {\tt $f_{\scst K}(\sigma):=0$,
      $\phi_{\scst K}(\sigma):=0$, for each $\sigma\in K\setminus \partial Q$; $H_{\scst K}:=H$.}\\
{\bf For } \= {\tt  $i=1$ to $i=m$  do}\\
 \>  {\tt take } \= {\tt a cell, $\sigma$,  of $f_{\scst K}\partial(\sigma_i)$, then}\\
\>\>  $ H_{\scst_K}:=H\setminus \{\sigma\}$,  \\
\>\> {\bf For } \= {\tt  $k=1$ to $k=i-1$} {\tt  do}\\
\>\>\> $f_{\scst K}(\sigma_k):= f_{\scst K}(\sigma_k)+\langle \sigma, f_{\scst K}(\sigma_k)\rangle f_{\scst K}\partial(\sigma_i)$\\
\>\>\> $\phi_{\scst K}(\sigma_k):= \phi_{\scst K}(\sigma_k)+\langle \sigma, f_{\scst K}(\sigma_k)\rangle(\sigma_i+ \phi_{\scst K}\partial(\sigma_i))$\\
 {\sc Output:} {\tt the AT-model $(K,H_{\scst K},f_{\scst K}, g_{\scst K}, \phi_{\scst K})$ for
$K$.}
\end{tabbing}
\end{algorithm}

Observe that in the algorithm, a cycle is never created because the cycles of $\partial Q$ are also cycles of $K$. Therefore, when a cell $\sigma_i$ of $K$ is added, then 
$f_{\scst K}\partial(\sigma_i)$ is never null and therefore
 a class of homology is always eliminated (that is, a cell $\sigma$ of $f_{\scst K}\partial(\sigma_i)$ is removed from $H_{\scst K}$). 
Alg. \ref{alghom} is ${\cal O}(m^3)$, where $m$ is the number of cells of $K\setminus \partial Q$ (worst case complexity).

\begin{figure}
\centering
\includegraphics[width=12cm]{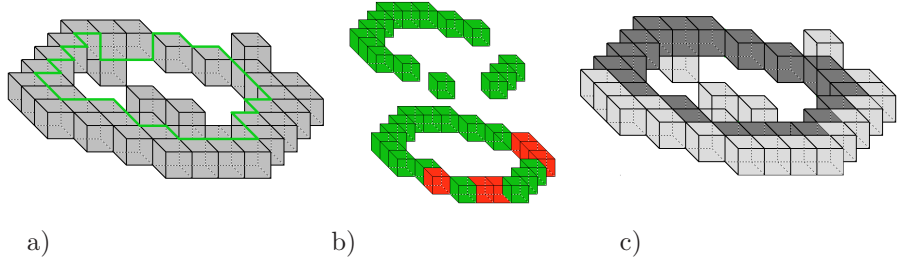}\\
\begin{footnotesize} a) \hspace{3.5cm} b)\hspace{3.5cm}c)\hspace{3cm}$\mbox{ }$\end{footnotesize}
\caption{a) A cubical complex $\partial Q$ and a representative cycle of a homology
generator; b) in green, voxels  considered the first time the edges of the cycle are visited. In red, voxels considered in the second visit;
c) in black, voxels  representing the cycle. }
\end{figure}

\subsection{From Cubical Complexes to Digital Pictures}

Now, given a digital picture $I$, suppose that we have computed an AT-model $(K,H_{\scst K},f_{\scst K},g_{\scst K},\phi_{\scst K})$ for $K$, following the steps given in Subsections \ref{picture}, \ref{partialq} and \ref{q}.
For each $\sigma$ in $H_{\scst K}$, $g_{\scst K}(\sigma)$ is  a representative cycle of a homology generator of $K$ and, therefore, of $Q$, since the homology of $K$ and $Q$ coincide and the representative cycles are in $\partial Q$.
Recall that if $\sigma$ is a vertex, then $g_{\scst K}(\sigma)$ is a vertex  representing a connected component; if $\sigma$ is an edge, then 
$g_{\scst K}(\sigma)$ is a sum of edges  representing a tunnel; and if $\sigma$ is a square, then $g_{\scst K}(\sigma)$ is a sum of squares  representing a cavity.

Given a representative cycle $g_{\scst K}(\sigma)$ of a homology generator, our aim in this subsection is to draw the equivalent cycle in the picture $I$.  

First, if $g_{\scst K}(\sigma)$ is a vertex then,   $g_{\scst K}(\sigma)$ is a face of a square in $\partial Q$. This square is shared by a voxel $V$ of $B$ and a voxel of $\Z^3\setminus B$. Then, associate the voxel $V$ to the vertex  $g_{\scst K}(\sigma)$.

Second, if $g_{\scst K}(\sigma)$ is a sum of edges, suppose that $g_{\scst K}(\sigma)$ is a simple cycle (if not, it can always be decomposed in simple ones). Visit all the edges of the cycle in order. If an edge, $a$, and the next edge, $b$, are facets of a square $\sigma\in \partial Q$, then associate the single voxel $V$ of $B$ which has $\sigma$ in its boundary, to the edges $a$ and $b$. 
After that,  visit  all the  edges that have not been associated to any voxel. If a voxel $V$ is associated with the next edge of the current one, $a$,  and    there is a voxel $V'\in B$ having $a$ in its boundary, such that
$V'$ is in the boundary of $I$ and $V'$ and $V$ are  $6$-neighbor, then associate $V'$ to $a$. If not, look at the previous edge  and do the same procedure.

\begin{figure}[t]
\centering
\includegraphics[width=8cm]{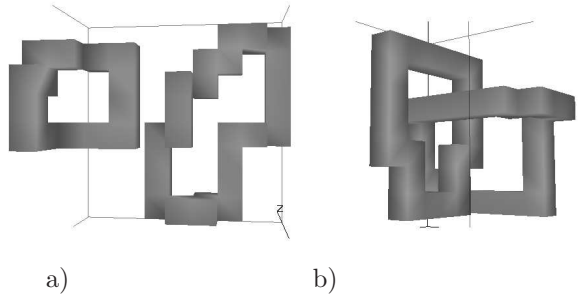}\\
\begin{footnotesize}a) \hspace{3cm} b)\hspace{3cm}\end{footnotesize}
\caption{a) Two non-linked circles; b) two once-linked circles.}\label{i12}\end{figure}

If not, take any voxel of $B$  that contains  $a$, having a $6$-neighbor voxel in $\Z^3\setminus B$.     

Finally, if $g_{\scst K}(\sigma)$ is a sum of squares,  associate, to each square $\sigma$, the single voxel  $V$ of $B$ which has $\sigma$ in its boundary.

\begin{figure}[t]
\includegraphics[width=10cm]{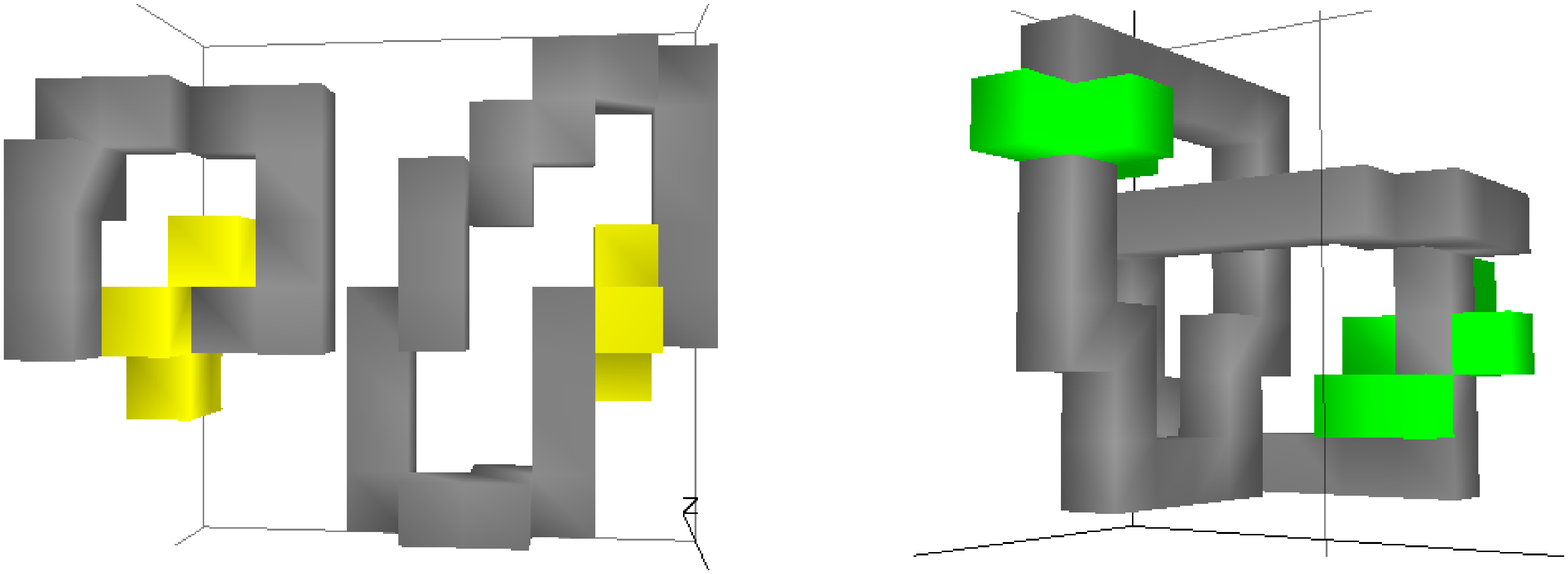}\\
 \begin{center}\begin{tabular}{|c|ccc|}\hline $Q'_1$&$\alpha_1\smile_{\scriptscriptstyle Q}\alpha_1$&$\alpha_1\smile_{\scriptscriptstyle Q} \alpha_2$&$\alpha_2\smile_{\scriptscriptstyle Q}\alpha_2$ \\\hline
$\beta_1$&$0$&$0$&$0$     \\   
$\beta_2$& $0$&$0$&$0$\\\hline
$Q'_2$&$\alpha'_1\smile_{\scriptscriptstyle Q}\alpha'_1$&$\alpha'_1\smile_{\scriptscriptstyle Q} \alpha'_2$&$\alpha'_2\smile_{\scriptscriptstyle Q}\alpha'_2$ \\\hline
$\beta'_1$&$0$&$1$&$0$        \\
$\beta'_2$& $0$&$1$&$0$\\
\hline
\end{tabular}\end{center}	\caption{ In yellow, representative cycles of the two tunnels of $Q'_1$; 
in green, representative cycles of the two tunnels of $Q'_2$; at the bottom,  the tables for the cup product on $Q'_1$ and $Q'_2$, where $\alpha_i$ (resp. $\alpha'_i$), $i=1,2$, are representative cycles of the two tunnels and $\beta_i$ (resp. $\beta'_i$), $i=1,2$, are representative cycles of the two cavities of $Q'_1$ (resp. $Q'_2$).}\label{aros}
\end{figure}

\subsection{Cohomology ring of $Q$}

Given a digital picture $(\Z^3,26,6,B)$, its associated cubical complex $Q$ and having computed an AT-model $(K,H_{\scst K},f_{\scst K},g_{\scst K},\phi_{\scst K})$ for $K$, the last step of the process is the  computation of the cohomology ring. This can be performed using the formula for the cubical cup product given in Def. \ref{cupcubico}.

\begin{example}
This example shows an application of $\smile_{\scriptscriptstyle Q}$ to discriminate different embeddings of the same object. Consider the cubical complex $Q_1$ (resp. $Q_2$) associated to a digital picture where the set $B$ consists in two once-linked ``circles'' (resp. two unlinked ``circles''). See Fig. \ref{i12}.
 Both complexes have two tunnels and no cavities, so these properties are not able to distinguish them. 

Now, denote by $Q'_1$ and $Q'_2$, the cubical complexes associated to the background of    $I_1$ and $I_2$ (white voxels of Fig. \ref{i12}). 
Compute an AT-model for $Q'_i$, and  its cohmology ring, for $i=1,2$. We obtain that the multiplication table for the cup product on $Q'_1$ is null whereas on $Q'_2$ is not (see Fig. \ref{aros}). This fact allows us to assert that the two complexes $Q'_1$ and $Q'_2$ are not topologically equivalent. 
\end{example}

\begin{figure}[t]
\centering
\includegraphics[width=8cm]{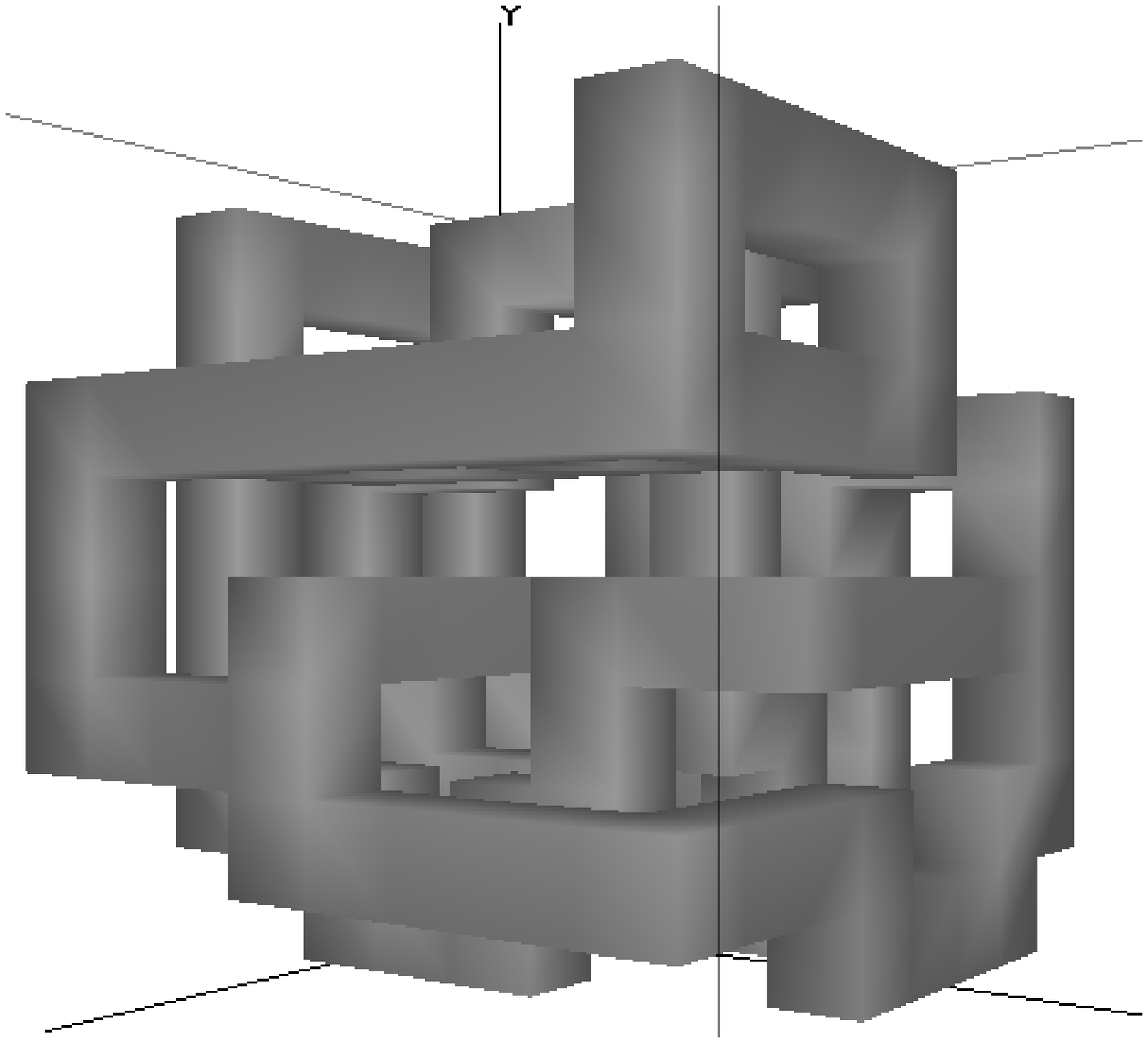}
\includegraphics[width=10cm]{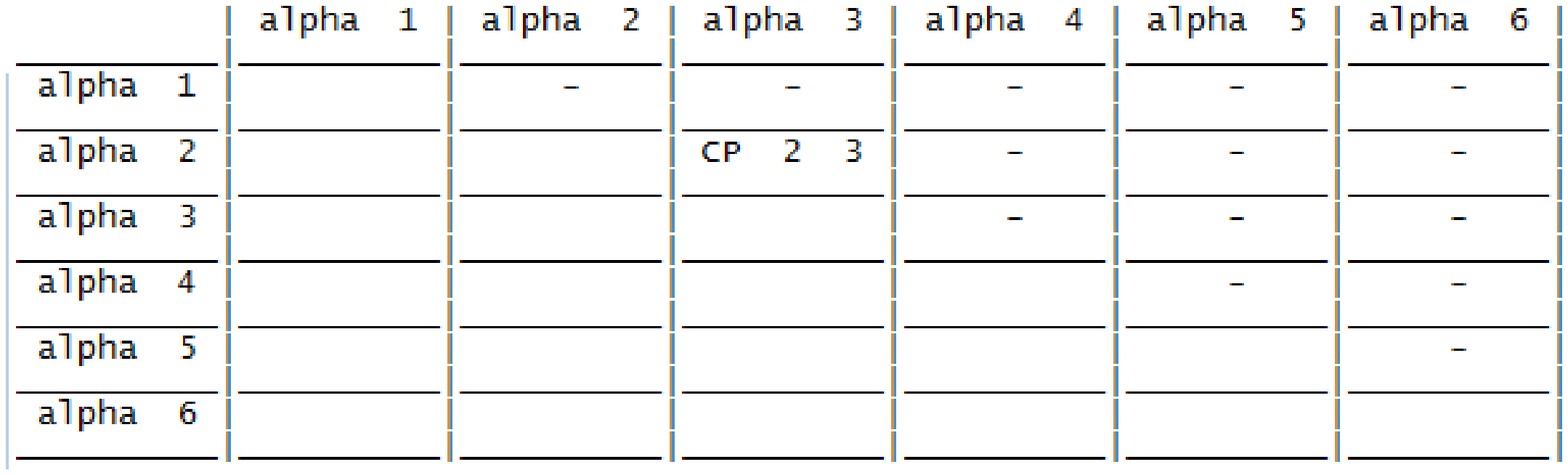}
\caption{A configuration of $6$ linked circles and the results of the computation of the cup product of the cubical complex associated to the white voxels of the picture.}\label{10}\end{figure}

\begin{figure}[t]
\centering
\includegraphics[width=8cm]{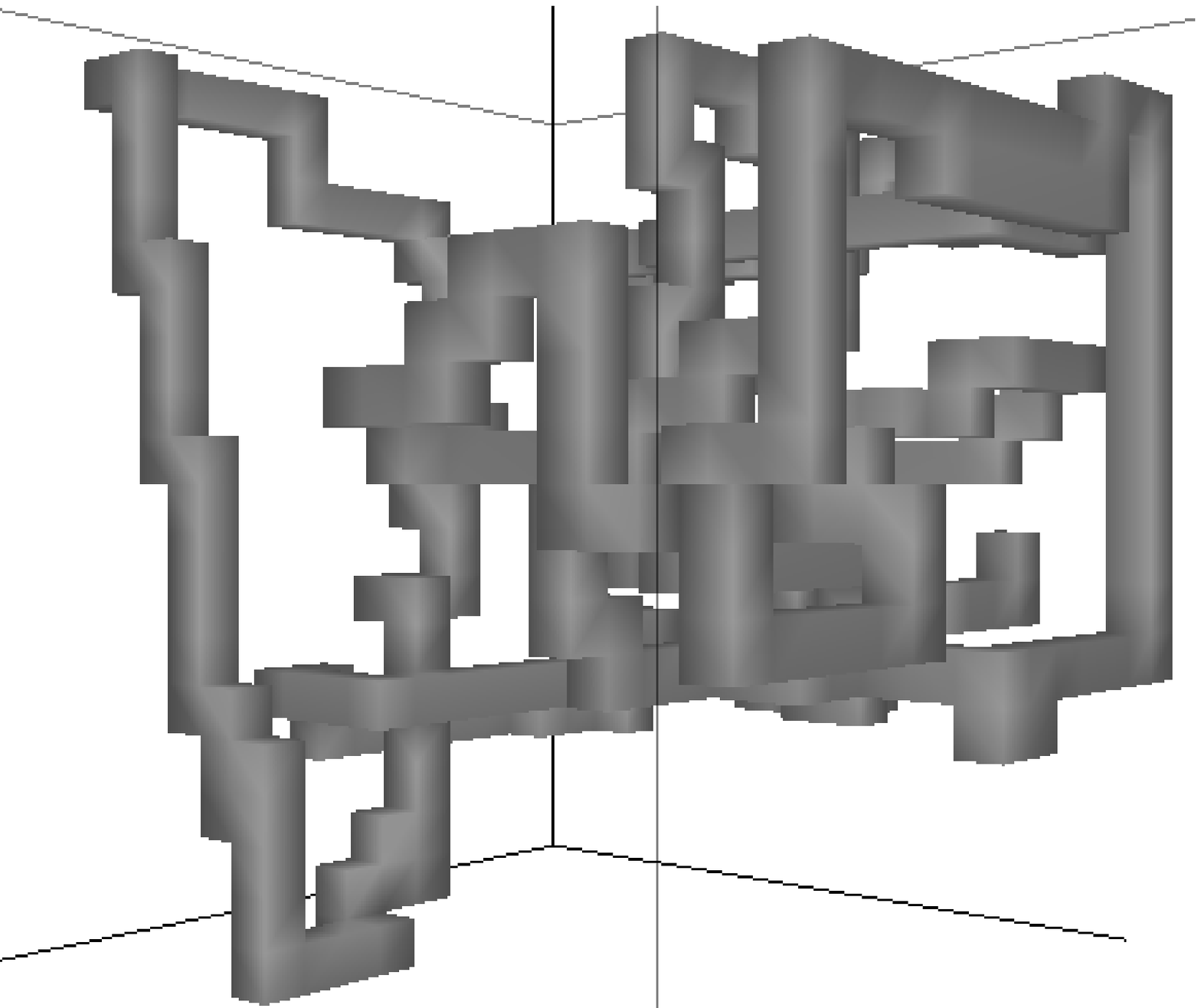}
\includegraphics[width=10cm]{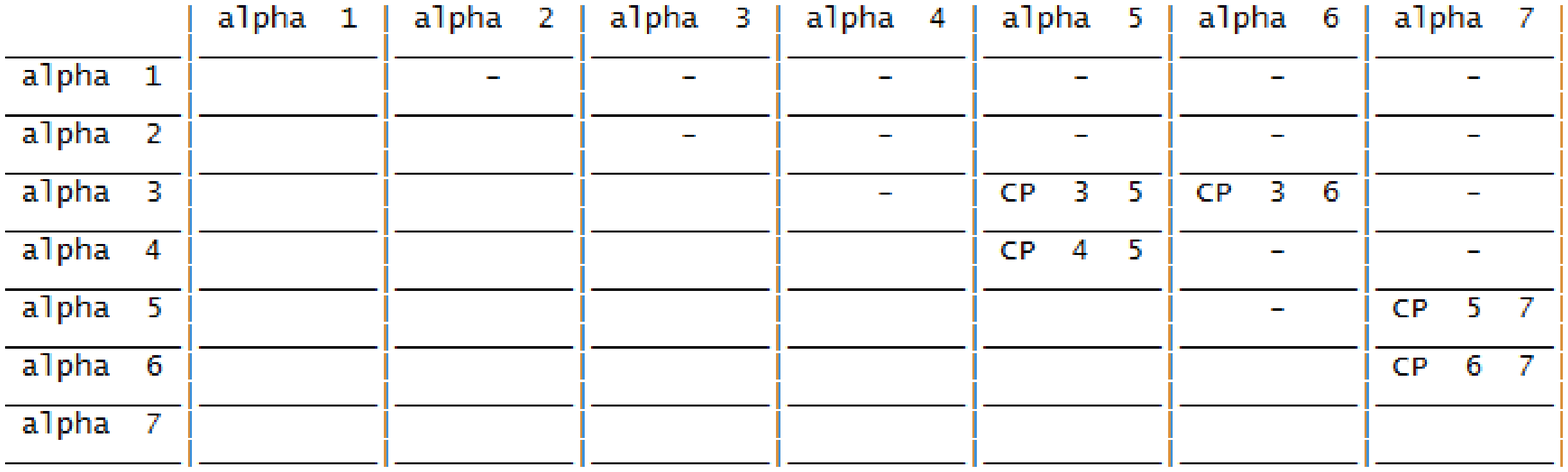}
\caption{A configuration of $7$ linked circles and the results of the computation of the cup product of the cubical complex associated to the white voxels of the picture.}\label{20}\end{figure}

\begin{example}
Consider the picture in Fig. \ref{10}. The cubical complex associated to the white voxels of the picture has $1$ connected component, $6$ tunnels and $3$ cavities. The cup product is trivial for any two pairs of homology classes except for two tunnels named as $\alpha_2$ and $\alpha_3$ wich is the sum of two of the three cavities (see Fig. \ref{10}).

Finally, Consider the picture in Fig. \ref{20}. The cubical complex associated to the white voxels of the picture has $1$ connected component, $7$ tunnels and $12$ cavities. The results of the computation of the cup product can be seen in the table of Fig. \ref{20}, where ``CP $i$ $j$" means the sum of the cavities $i$ and $j$. 
\end{example}


\begin{example}

MRA of chest showing the heart and great vessels.

\begin{figure}
\centering
\includegraphics[width=12cm]{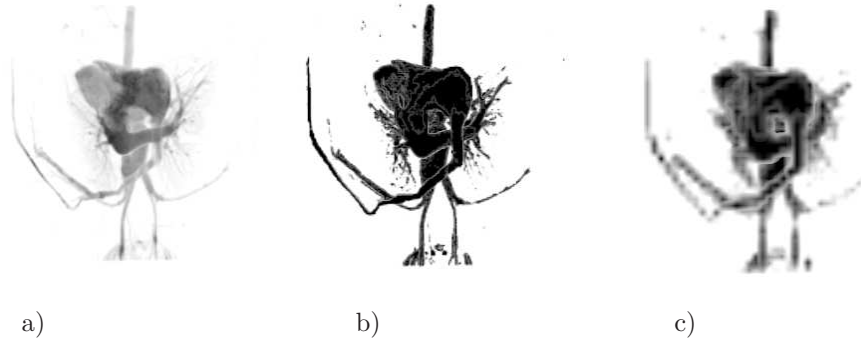}\\
\begin{footnotesize}a) \hspace{3.9cm} b)\hspace{3.9cm}c)$\mbox{ }$\hspace{2cm}\end{footnotesize}
\caption{a) Original; b) binarized; c) resized.}\label{real}
\end{figure}

Originally it was a stack of 14 gray-scale digital images in DICOM format, each one of size 320x320. This 2D stack can be see as the 3D digital image ``chest" of dimensions 320x320x14  (see Fig. \ref{real}.a). 
After a binary process, we obtain the picture $I$ in Fig \ref{real}.b).
``resized"  80x80x28

Let $Q$ be the cubical complex that represents $I$, let $\partial Q$ be 

$$\begin{array}{lc}
\\
\mbox{Number of black voxels of $I$} & 18,062
\\
\mbox{Number of cells of $Q$} & 179,769
\\
\mbox{Number of cells of $\partial Q$}& 66,143
\\
\mbox{Number of cell of $K$}
\\
\mbox{AT-model implementation execution time} 
\end{array}$$

\end{example}


\section{Conclusions and Future Work}\label{future}

In this paper we present formulas to directly compute the cohomology ring of 3D cubical complexes and develop a method for the computation on 3D binary-valued pictures. This computation on cubical complexes can be regarded as a starting point to compute the cup product on general polyhedral cell complexes, which is, in fact, our last goal. As related work, we must mention  \cite{MR08}, where homology of 3D pictures using particular cell structures provided by the 26-adjacency is performed. 
The restriction to the 3D-world allows to work over ${\bf Z}/2$, what facilitates the calculus. However, a harder task could be the one of extending the formulas of the cohomology ring to higher dimensions what could be applied to more general contexts out of digital images. In this sense, cohomology ring of nD simplicial complexes using AM-models and working in the integer domain, has been established in \cite{GJMR09}.
Another goal for future work is the one of applying theoretical results to irregular graph pyramids and compute the cohomology ring on the cell complexes associated to such structures. In the paper \cite{GIIK09}, representative cocycles for cohomology generators on irregular graph pyramids are computed, what can be considered a first step in this direction.



\begin{thebibliography}{4}
\bibitem{AH35} Alexandroff P., Hopf H.: Topologie I. Springer, Berlin 1935

\bibitem{GJM09} Gonzalez-Diaz R., Jimenez M.J., Medrano B.: Cohomology ring of 3D cubical complexes. Proc. of the 13th. International Workshop on Combinatorial Image Analysis, IWCIA 2009. Progress in Combinatorial Image Analysis (2009) 139-150.

\bibitem{GJMMR08} Gonzalez-Diaz R., Jimenez M.J., Medrano B., Molina-Abril H., Real P.:
Integral Operators for Computing Homology Generators at Any Dimension.
Proc. of the 13th Iberoamerican Congress on Pattern Recognition, CIARP 2008. LNCS v. 5197, 356-363 (2008)


\bibitem{GR03}Gonzalez-Diaz R., Real P.: Towards Digital
Cohomology. Proc. of the 11th Int. Conf. on Discrete Geometry for Computer Imagery,
               DGCI 2003, LNCS, v. 2886 (2003) 92-101


\bibitem{GR05} Gonzalez-Diaz R., Real P.: On the Cohomology of $3D$
Digital Images. Discrete Applied Math., v. 147 (2005) 245-263


\bibitem{GJMR09} Gonzalez-Diaz R., Jimenez M.J., Medrano B., Real P.: Chain homotopies for object topological representations. Discrete Applied Mathematics, v. 157 (2009) 490-499


\bibitem{GIIK09} Gonzalez-Diaz R., Ion A., Iglesias-Ham, M. and Kropatsch, W. G.:
Irregular Graph Pyramids and Representative Cocycles of Cohomology Generators. Int. Workshop GbRPR09. LNCS v. 5534 (2009) 263-272

\bibitem{MR08} Molina-Abril H., Real P.: Advanced Homological Information of Digital Volume via Cell Complexes. Int. Workshop SSPR 2008, LNCS v. 5342 (2008) 361-371

\bibitem{KMM04} Kaczynski T., Mischaikow K.,  Mrozek M.:
 Computational Homology. Applied Mathematical Sciences, v. 157,
 2004

\bibitem{Kad98} Kadeishvili, T.:  DG Hopf Algebras with Steenrods i-th coproducts. Bull. Georgian Acad. Sci. 158 (1998) 203-206



\bibitem{Mun84} Munkres J.R.: Elements of Algebraic
Topology. Addison-Wesley  Co. 1984


\bibitem{vessel} Niethammer, M., Stein, A.N., Kalies, W.D., Pilarczyk, P., Mischaikow, K., Tannenbaum, A.: Analysis of blood vessel topology by cubical homology.   2002 International Conference on Image Processing, Proceedings IEEE ICIP, v. 2 (2002) 969-972


\bibitem{PIKDH08} Peltier S., Ion A., Kropatsch W.G., Damiand G., Haxhimusa Y.:
Directly computing the generators of image homology using graph pyramids.
Image and Vision Computing, In Press, Corrected Proof.

\bibitem{Ser51} Serre J.P.: Homologie Singuliere des espaces fibres, applications. Ann. Math. v. 54 (1951) 429-501


\end{thebibliography}
\end{document}